%% file: main.tex
\documentclass{article}

\usepackage[preprint,nonatbib]{neurips_2024}

\usepackage[numbers]{natbib}

\usepackage{wrapfig,stfloats}
\usepackage{subcaption}
\usepackage[utf8]{inputenc} 
\usepackage[T1]{fontenc}    
\usepackage{hyperref}       
\usepackage{url}            
\usepackage{booktabs}       
\usepackage{amsfonts}       
\usepackage{nicefrac}       
\usepackage{microtype}      
\usepackage{xcolor}         
\usepackage{graphicx}
\usepackage{xspace}
\usepackage{multirow}
\usepackage{tikz}
\usepackage{comment}
\usepackage{enumitem}

\newcommand{\incr}[1]{\textcolor{blue}{({#1}\textbf{$\uparrow$})}}
\newcommand{\decr}[1]{\textcolor{red}{({#1}\textbf{$\downarrow$})}}

\usepackage{pifont}

\begin{document}

\title{Does Chain-of-Thought Reasoning Help Mobile GUI Agent? An Empirical Study}

\author{
Li Zhang\thanks{Authors contributed equally to this work.} \quad Longxi Gao$^{*}$ \quad Mengwei Xu \\
Beijing University of Posts and Telecommunications \\
\texttt{\{li.zhang,gaolongxi,mwx\}@bupt.edu.cn}
}

\maketitle
\input{abstract.tex}
\input{intro.tex}
\input{bkgnd.tex}
\input{design.tex}
\input{eval.tex}

\input{conclusion.tex}

\bibliographystyle{abbrvnat}
\bibliography{agents-ref}
\appendix\onecolumn
\input{appendix.tex}

\end{document}

%% file: abstract.tex
\begin{abstract}
Reasoning capabilities have significantly improved the performance of vision-language models (VLMs) in domains such as mathematical problem-solving, coding, and visual question-answering.
However, their impact on real-world applications remains unclear.
This paper presents the first empirical study on the effectiveness of reasoning-enabled VLMs in mobile GUI agents, a domain that requires interpreting complex screen layouts, understanding user instructions, and executing multi-turn interactions.
We evaluate two pairs of commercial models--Gemini 2.0 Flash and Claude 3.7 Sonnet--comparing their base and reasoning-enhanced versions across two static benchmarks (ScreenSpot and AndroidControl) and one interactive environment (AndroidWorld).
We surprisingly find the Claude 3.7 Sonnet reasoning model achieves state-of-the-art performance on AndroidWorld. 
However, reasoning VLMs generally offer marginal improvements over non-reasoning models on static benchmarks and even degrade performance in some agent setups.
Notably, reasoning and non-reasoning VLMs fail on different sets of tasks, suggesting that reasoning does have an impact, but its benefits and drawbacks counterbalance each other.
We attribute these inconsistencies to the limitations of benchmarks and VLMs.
Based on the findings, we provide insights for further enhancing mobile GUI agents in terms of benchmarks, VLMs, and their adaptability in dynamically invoking reasoning VLMs.
The experimental data are publicly available at \url{https://github.com/LlamaTouch/VLM-Reasoning-Traces}.
\end{abstract}

%% file: intro.tex
\section{Introduction}

The reasoning capabilities significantly enhance large language models (LLMs) and vision-language models (VLMs) by utilizing long chain-of-thought (CoT) thinking and extended test-time computation~\cite{snell2024scaling,wei2022chain}.
Empirical evidence from recent studies demonstrates that such enhanced reasoning abilities yield superior performance in domains like mathematical problem-solving, coding, and visual question answering~\cite{snell2024scaling,guo2025deepseek,wang2024exploring}.
These models with reasoning capabilities have established new benchmark records in their respective fields, surpassing previous LLMs/VLMs that lack reasoning.

Despite these advancements, the complexities inherent in real-world applications pose significant challenges.
\textit{Does reasoning help real-world complex multimodal tasks, beyond coding and math?}
In this study, we focus on a practical, unsolved task, a.k.a. mobile GUI agents, particularly for mobile device control tasks~\cite{wen2023empowering,rawles2023android,rawles2024androidworld,wen2024autodroid,xu2025every}, which present a unique testbed due to their intricate visual layouts, diverse functionalities, and the requirement for multi-step reasoning and interaction to achieve user goals. 
Existing state-of-the-art (SOTA) mobile GUI agents without reasoning still struggle to deliver satisfactory and practical success rates in real-world environments~\cite{wen2024autodroid,rawles2024androidworld,simularAgent}.
We hypothesize that incorporating reasoning ability, similar to its application in other domains, could potentially enhance the performance of mobile GUI agents by improving task comprehension, environmental adaptation, and action decision-making.
Therefore, evaluating the effectiveness of reasoning VLMs in this demanding downstream task is of critical importance.

\textbf{Methodology and Experiments.}
This study fills the existing gap by conducting a comprehensive empirical evaluation of reasoning VLMs in mobile GUI agents.
Specifically, we select two pairs of commercial models, Gemini 2.0 Flash~\cite{gemini-flash-thinking} and Claude 3.7 Sonnet~\cite{claude3.7}, both with and without reasoning capability (referred to as Gemini/Claude and Gemini/Claude Thinking, respectively).
Additionally, we take GPT-4o~\cite{gpt4o} without reasoning capability as a performance reference.
We select the following benchmarks\footnote{We are experimenting with more benchmarks.}.
\begin{itemize}
    \item Static benchmarks -- AndroidControl~\cite{li2024effects} and ScreenSpot~\cite{cheng2024seeclick}.
    \item Interactive testbed -- AndroidWorld~\cite{rawles2024androidworld}.
\end{itemize}
For each benchmark, we implement and test different agent setups upon the VLMs.

\textbf{Results and Findings.}
Through experiments and analysis, we make the following key observations.

(1) On static benchmarks, reasoning VLMs generally have marginal improvements over non-reasoning VLMs, and even suffer severe performance degradation under certain agent setups.
For instance, in AndroidControl, Gemini Thinking achieves an average action prediction accuracy of 54.4\%, only 0.8\% higher than the non-reasoning version.
In grounding tasks within ScreenSpot, performance improvements are observed only with Claude Thinking with normalized center-point output; in other setups, accuracy drops by up to 29.7\%.

(2) On the interactive mobile testbed AndroidWorld, Claude Thinking achieves a 64.7\% task completion rate with set-of-mark prompting, setting a SOTA record compared to the numbers reported in prior arts, and is 6.3\% higher than the non-reasoning version.
This highlights the effectiveness and potential of reasoning VLMs in real-world mobile GUI automation tasks.
Nonetheless, Gemini Thinking exhibits a slight performance drop compared to its base variant, indicating that improvements are model-specific.

(3) Surprisingly, the reasoning and non-reasoning VLMs fail on a substantially different set of test cases.
For example, Gemini Thinking fails on 36\%, 9\%, and 12\% of tasks in ScreenSpot, AndroidControl, and AndroidWorld, respectively, that Gemini can successfully accomplish.
Vice versa, Gemini Thinking also succeeds up to 10\% of tasks that Gemini fails.
This suggests that the lack of accuracy improvement at the benchmark level is not because reasoning has no effect, but rather its positive and negative impacts counterbalance each other.
These inconsistencies emphasize the need for a deeper investigation into the role of reasoning in mobile GUI agents.

(4) Our manual investigation of the reasoning process reveals that errors in reasoning VLMs stem from limitations in both mobile GUI agent benchmarks and the underlying VLMs.
We find that reasoning VLMs exhibit similarities to human thought processes when operating smartphones.
However, this advanced understanding does not translate into performance gains due to inherent benchmark limitations, such as vague task instructions and the inability to evaluate multiple possible actions within static datasets.
Furthermore, during the reasoning phase, VLMs sometimes fail to comprehend screen details accurately and may generate responses that are inconsistent with the reasoning processes.

(5) Reasoning VLMs significantly increase model output tokens by at least 3.11$\times$ and up to 14.78$\times$, leading to higher response latency and monetary costs without clear performance benefits.
As observed in ScreenSpot, the average number of output tokens increases from 37.6 to 238.5.
Without strict output constraints, reasoning VLMs may generate additional tokens in their final responses, e.g., to summarize their thought process.
This raises costs and practicality concerns regarding the indiscriminate use of reasoning VLMs for all mobile GUI agent tasks.

\textbf{Implications.} 
We derive several implications for enhancing mobile GUI agents by fully unleashing the reasoning capabilities of VLMs.
(1) Mobile GUI agents with reasoning VLMs are better to be evaluated on interactive testbeds, instead of static benchmarks.
This could avoid the intrinsic limitations of static benchmarks.
(2) The underlying VLMs should be specifically trained for mobile GUI agents to improve grounding and screen comprehension at the reasoning phase.
It is also crucial to maintain consistency from reasoning to final outputs.
(3) Resource efficiency~\cite{xu2025resource} will become a major obstacle toward reasoning-enhanced mobile GUI agent, due to the excessive task completion latency and token expense. Efficient reasoning is critical to a practical reasoning-enhanced mobile GUI agent.

\textbf{Contributions.} The contributions of this study are summarized as follows.
(1) We conduct the first empirical study of VLMs' reasoning capabilities in mobile GUI agents, a critical downstream task focused on automatic smartphone control.
(2) We demonstrate the limited performance gains from reasoning VLMs and highlight their limitations, particularly in failing tasks that non-reasoning VLMs can successfully complete.
(3) We perform an in-depth error analysis of the reasoning process, categorizing errors based on VLM limitations and benchmark constraints.
Our findings provide valuable insights for advancing future research in this area.
(4) We open-source the data, including the reasoning processes of VLMs, at \url{https://github.com/LlamaTouch/VLM-Reasoning-Traces}.

%% file: bkgnd.tex
\section{Background}

\subsection{Mobile GUI Agents}

\textbf{From API-based agents to GUI agents.}
Traditional mobile agents, like Apple Siri~\cite{siri} and Google Assistant~\cite{google-assistant}, relied on static, API-driven interactions. These agents operated based on predefined rules and could only automate tasks with exposed APIs, therefore limiting their adaptability.
Recently, leveraging the advancements in LLMs and VLMs, modern mobile GUI agents have shifted from API-dependent automation to direct interpretation and operation on mobile screens~\cite{wen2023empowering,wang2023enabling,li2024personal,openaiOperator,claudeComputerUse,zhang2024guiagentsurvey}.
Instead of being restricted by predefined API calls, these agents analyze screen contents, user instructions, and execute actions based on visual and textual information, making them more adaptable to various unseen tasks and applications~\cite{zhang2025apiagentsvsgui}.

\textbf{Limitations of current mobile GUI agents.}
Prior studies have highlighted the challenges of automating mobile GUI tasks, particularly in real-world settings~\cite{rawles2024androidworld,zhang2024llamatouch}.
Unlike API-based agents that operate on structured interfaces, GUI agents must interpret diverse and evolving screen layouts, extract relevant information, and execute actions.
This complexity leads to inconsistencies, as existing models struggle with intricate UI hierarchies, ambiguous elements, and dynamic content.
Moreover, mobile GUI agents have yet to fully capitalize on recent advancements in LLMs/VLMs, particularly their reasoning capabilities.
While these models excel in tasks such as mathematical problem-solving, programming, and visual question answering, their potential for reasoning in mobile GUI automation remains largely unexplored.
This study investigates whether integrating reasoning from LLMs/VLMs can enhance task completion in complex, dynamic, and previously unseen mobile environments.

\subsection{Reasoning LLMs/VLMs}

\textbf{LLMs/VLMs with reasoning capabilities.}
To enhance reasoning capabilities for solving more complex tasks, OpenAI's o1 series models~\cite{openai-o1} became the first commercial models to adopt the test-time scaling technique in 2024.
These models follow a ``think first, then answer'' approach.
Specifically, during the thinking stage, they process user requests and generate a detailed CoT~\cite{wei2022chain} for self-reflection and reasoning.
By allocating more computation to this stage, these models produce more accurate final answers, as demonstrated by their performance in solving complex mathematical, coding, and multimodal reasoning problems~\cite{wang2024exploring,guo2025deepseek,snell2024scaling}.
As a result, commercial LLMs/VLMs with strong reasoning capabilities have emerged to address textual and visual tasks, including DeepSeek-R1~\cite{guo2025deepseek}, Gemini 2.0 Flash Thinking~\cite{gemini-flash-thinking}, Claude 3.7 Sonnet~\cite{claude3.7}, and Grok 3 Beta~\cite{grok3beta}, etc.

\input{fig-reasoning-example.tex}

We deem \textbf{multimodal reasoning is essential for mobile GUI agents}, as they encounter unseen and complex tasks within mobile apps.
These complexities increase in dynamic and unpredictable mobile contexts, such as frequently changing app content and intermittent network conditions.
Allocating time for reasoning allows mobile GUI agents to adapt to environmental changes, correct their own mistakes, and ultimately determine the optimal path for completing a GUI automation task~\cite{claude-extended-thinking}.
Figure~\ref{fig:reasoning-example} illustrates a reasoning process using Gemini 2.0 Flash Thinking in mobile GUI automation tasks.
In this example, the app is currently on the Message tab within the Discord app.
Given the task ``Set my DM Spam filter to `Do not filter direct messages' on Discord app'', the model initiates the reasoning process by explicitly outputting the task instruction and describing the current screen state in natural language.
It then analyzes the relationship between the task instruction and all visible GUI elements to identify the most relevant element and action—``clicking the tab named You'' that may help complete the task.
After selecting an action, the model reflects on its goal, the current GUI state, and the chosen action to further validate its decision.
Finally, it confirms its decision and clearly outputs the selected action in its response.
Based on this reasoning process, we observe that mobile GUI agents develop a comprehensive understanding of both the current GUI state and the task instruction, leading to confident and accurate actions.
This deeper understanding is crucial for handling complex and dynamic mobile environments.

Despite these promising outcomes, very few studies have explored how such reasoning processes benefit mobile GUI agents~\cite{zhan2023you,zhang2024android}.
Existing research primarily uses CoT prompting in highly controlled environments.
This study aims to bridge this gap by conducting a large-scale, comprehensive investigation into whether reasoning improves mobile GUI agent performance in real-world scenarios using VLMs with intrinsic reasoning capabilities.

%% file: fig-reasoning-example.tex
\begin{figure}[t]
	\centering
	\includegraphics[width=1\textwidth]{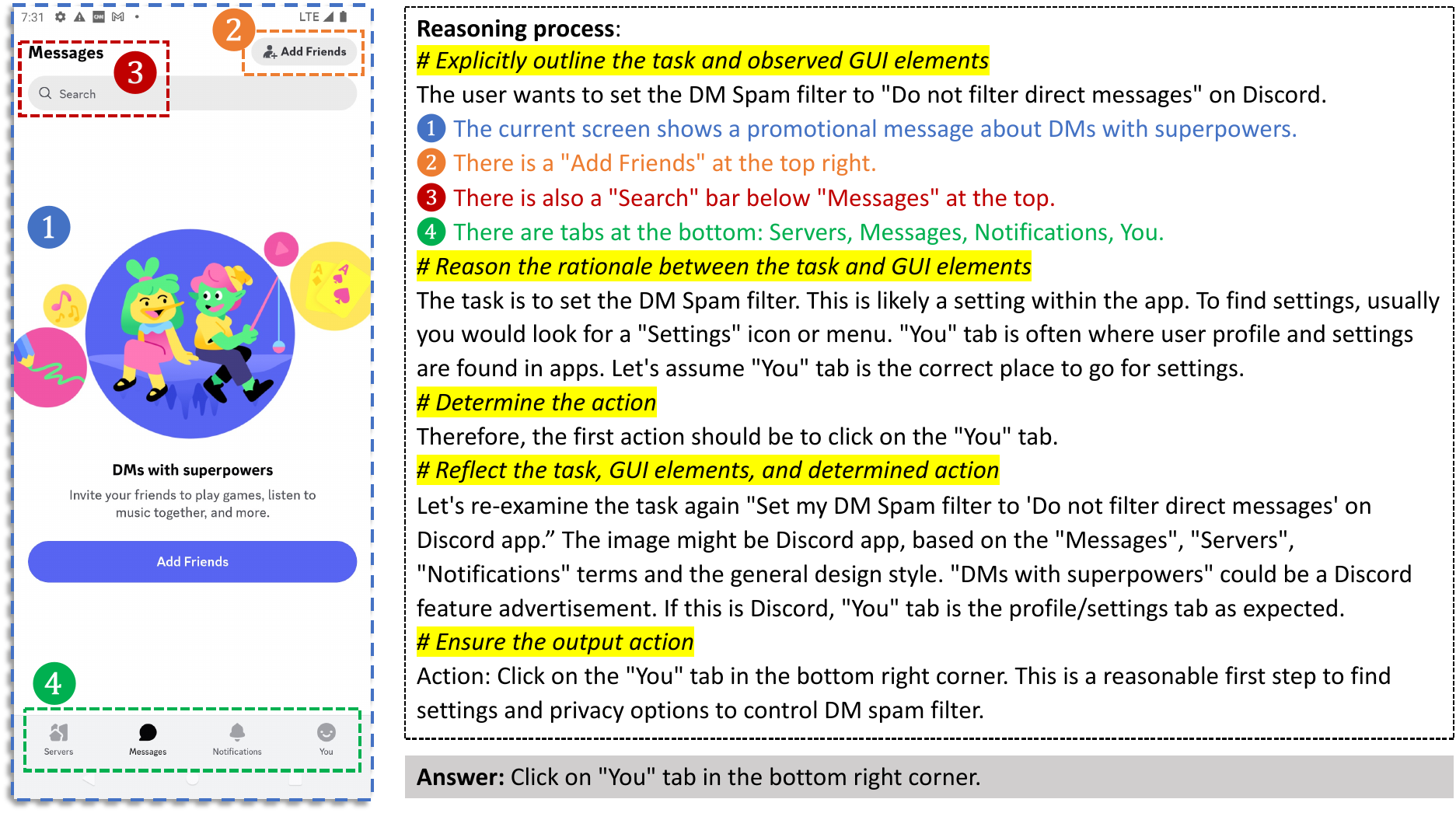}
	\caption{A demonstration of Gemini 2.0 Flash Thinking's reasoning process for mobile GUI automation tasks.
    The model first explicitly outlines the task instruction and the observed GUI elements, then reasons through the information to determine the actions.
    User request to the mobile GUI agent: \textit{You need to complete the task ``Set my DM Spam filter to `Do not filter direct messages' on Discord'', output possible actions on this GUI that may complete the task.}
    Left: The input mobile GUI (screenshot).
    Right: VLM's reasoning process and final response (action).}
	\label{fig:reasoning-example}
\end{figure}

%% file: design.tex
\section{Methodology}
\label{sec:method}

In this section, we describe the methodology employed in this empirical study.
First, we introduce the selected benchmarks and explain the rationale behind their selection.
Next, we detail the VLMs with and without reasoning capabilities, along with the mobile GUI agents built on top of them.
Finally, we outline the metrics used to evaluate their performance.

\textbf{Benchmarks.}
It is crucial to carefully select benchmarks for evaluating mobile GUI agents.
Prior studies have made extensive efforts to test mobile GUI agents using static datasets~\cite{rawles2024androidworld,li2024effects,sun2022meta,cheng2024seeclick,deng2024mobile}, but these approaches have proven inefficient in handling real-world mobile environments~\cite{rawles2024androidworld,zhang2024llamatouch}.
In this study, we incorporate both representative static and interactive benchmarks as follows.

\textit{(1) ScreenSpot~\cite{cheng2024seeclick}} is a GUI grounding dataset with more than 600 GUIs and over 1.2K task instructions, which is designed to assess the basic grounding capability of VLMs.
A grounding task is defined as: given a task instruction, the VLM identifies the corresponding GUI component to be acted upon, and outputs its coordinates.
This benchmark aims to reveal whether reasoning enhances the basic grounding capability of mobile GUI agents.
In our experiments, we use the ``mobile'' subset within ScreenSpot.

\textit{(2) AndroidControl~\cite{li2024effects}} is a static dataset for training and evaluating mobile GUI agents.
It is proposed by Google and contains more than 14K tasks across 800+ Android apps.
A key distinction from previous static datasets is its high-quality task annotations, comprehensive GUI representations, and inclusion of single-step task instructions, which facilitate the evaluation of different VLM prompting strategies.
In this study, we follow the experimental setup and evaluation approach used in AndroidControl and randomly select 500 tasks to approximate the results of the full test split.

\textit{(3) AndroidWorld~\cite{rawles2024androidworld}} is an interactive mobile GUI agent benchmark proposed by Google.
It uses predefined function calls to access internal app states for task completion verification, enabling a more accurate evaluation.
We incorporate AndroidWorld to assess existing mobile GUI agents across its 116 tasks, demonstrating their capabilities in real-world scenarios.

\textbf{Models and Agents.}
Models combined with curated prompts form mobile GUI agents.
We use two pairs of VLMs--Gemini 2.0 Flash~\cite{gemini-flash-thinking} and Claude 3.7 Sonnet~\cite{claude3.7}--each including a base model without reasoning and its reasoning-enabled variant\footnote{Gemini base model: \textit{gemini-2.0-flash-001}, reasoning model: \textit{gemini-2.0-flash-thinking-exp-01-21}. Claude base model: \textit{claude-3-7-sonnet-20250219}, reasoning model: \textit{claude-3-7-sonnet-20250219} with thinking mode enabled and a budget token number of 1024.}.
Additionally, we use GPT-4o~\cite{gpt4o}, which lacks reasoning capability, as a performance reference.
The mobile GUI agents in this study are built on top of these VLMs but differ in their prompting designs.
We primarily utilize agents released or open-sourced in prior studies.
For ScreenSpot, we instruct the agent to output three different formats for a grounded GUI element:
(1) normalized bounding box \textit{(e.g., [0.08, 0.688, 0.92, 0.735])}; (2) pixel-based bounding box \textit{(e.g., [127, 34, 235, 978])}; and (3) normalized center point \textit{(e.g., [255, 370])}.
For AndroidControl, we use the ER prompt, which takes the task instruction and previous action list as input.
We further modify its input to get three variants: (1) task instruction only; (2) step instruction only; and (3) task and step instructions.
For AndroidWorld, we employ three agent designs:
(1) M3A with set-of-mark prompting~\cite{yang2023set}; (2) M3A with accessibility tree (a11y tree) prompting; and (3) T3A with a11y tree prompting.

\textbf{Metrics.}
On static mobile GUI benchmarks, we report grounding accuracy for ScreenSpot and action prediction accuracy for AndroidControl.
The evaluation method for AndroidControl follows the approach detailed in its original work ~\cite{li2024effects}.
For AndroidWorld, we assess end-to-end task completion rates.
During experiments, we log all traces, including model responses, reasoning processes, and screenshots, for token count and error analysis.

%% file: eval.tex
\section{Experimental Results}
\label{sec:exp}

In this evaluation, we examine the performance of mobile GUI agents with and without integrated reasoning capabilities.
First, we report task completion accuracies across all tasks at the benchmark level (\S\ref{sec:eval-static}).
Next, we analyze individual tasks to determine whether the reasoning process benefits GUI agents by distinguishing task completion status (\S\ref{sec:eval-task-analysis}).
Then, we categorize errors introduced during the reasoning process (\S\ref{sec:err-analysis}).
We also provide a comparative token count analysis to show the cost of utilizing reasoning VLMs (\S\ref{sec:token-cost}).
Finally, we derive key implications for further enhancing reasoning-enabled mobile GUI agents (\S\ref{sec:implications}).

\input{tab-acc-screenspot}
\input{tab-acc-androidcontrol}

\subsection{Benchmark-level Analysis}
\label{sec:eval-static}

\textbf{Static benchmarks.}
The results presented in Table~\ref{tab:screenspot} and~\ref{tab:android-control} reveal a trend: \textit{reasoning VLMs generally do not improve the performance of mobile GUI agents on static benchmarks.}
In some agent setups, it even leads to a significant performance drop.

Specifically, in ScreenSpot~\cite{cheng2024seeclick}, we evaluate GUI grounding accuracy across different VLMs and grounding output formats.
As shown in Table~\ref{tab:screenspot}, reasoning generally degrades grounding accuracy when using normalized and pixel-based bounding boxes across VLMs.
Gemini Thinking and Claude Thinking exhibit substantial accuracy reductions (28.7\% and 16.1\%, respectively, for normalized bounding boxes), indicating a negative impact on this grounding task.
However, for normalized center points, the effect of reasoning is mixed: while Gemini Thinking's accuracy significantly declines (29.7\%), Claude Thinking improves by 8.6\%.
This highlights that the effectiveness of reasoning is highly model-dependent and task-specific, with potential benefits for precise center-point localization in Claude Thinking.
In AndroidControl, as shown in Table~\ref{tab:android-control}, reasoning VLMs provide only a marginal improvement in accuracy across VLMs and agent designs.
Gemini Thinking increases accuracy by an average of just 0.75\%, while Claude Thinking sees a slight improvement of 2.3\%.
Overall, our evaluation across two distinct static GUI benchmarks suggests that integrating reasoning VLMs into mobile GUI agents does not consistently improve performance and, under some setups, may even hinder their effectiveness.

\textbf{Interactive testbed.}
We then use AndroidWorld as an interactive testbed to evaluate mobile GUI agents, along with three different agent setups proposed in their study~\cite{rawles2024androidworld}.
The results in Table~\ref{tab:android-world} indicate that different model pairs exhibit distinct behaviors.
With reasoning enabled in Gemini, performance drops by an average of 2.7\%, demonstrating its non-positive impact on task completion rates.
In contrast, Claude Thinking enhances the performance with an average improvement of 6.3\%.
Surprisingly, task completion rates increase by up to 9.5\% in the M3A-SoM setup with reasoning enabled, achieving SOTA performance on AndroidWorld.
\input{tab-acc-androidworld}

We further analyze task completion rates categorized by difficulty levels in AndroidWorld.
The results are shown in Figure~\ref{fig:androidworld-difficulty}.
The key observation is that Claude Thinking solely improves task completion rates over its base model on easy and medium tasks while delivering nearly identical performance on hard tasks.
This suggests that, at present, reasoning VLMs still fall short in handling complex interactive tasks, indicating that they are not a silver bullet for generalized mobile GUI agent tasks.

\input{fig-androidworld-difficulty}
\input{tab-task-stats}

\subsection{Task-level Analysis}
\label{sec:eval-task-analysis}

From the prior results, we conclude that reasoning VLMs do not benefit mobile GUI agents in static benchmarks, as they typically achieve comparable performance regardless of whether reasoning is enabled.
Their improvement in AndroidWorld is model-specific but not substantial.
Moreover, the reported performance is based on the entire dataset, without assessing the impact of reasoning capabilities on individual tasks.
In this section, we conduct a deeper analysis of individual tasks within each benchmark to determine whether reasoning VLMs enhance or hinder mobile GUI agent performance.

We focus on two categories of tasks.
(1) Tasks that cannot be completed by non-reasoning models but are successfully completed with reasoning VLMs (\textbf{F$\rightarrow$T} in Table~\ref{tab:task-stats}).
These tasks demonstrate the advantages of reasoning VLMs in improving mobile GUI agents.
(2) Tasks that can be completed with non-reasoning VLMs but fail in reasoning VLMs (\textbf{T$\rightarrow$F} in Table~\ref{tab:task-stats}).
These tasks highlight potential limitations of current reasoning VLMs, which may halt their integration into existing mobile GUI agents.

\textbf{Observation: Result inconsistency after the adoption of reasoning VLMs.}
Our results are presented in Table~\ref{tab:task-stats} with the following observations.
First, applying reasoning to previously successful tasks introduces a substantial number of inconsistencies, which undermines the accuracy achieved by mobile GUI agents in non-reasoning mode across most benchmarks and experimental setups.
For example, on ScreenSpot with normalized bounding-box output, Gemini and Claude fail 36\% and 18\% of tasks after reasoning, respectively, even though having successfully completed these tasks without reasoning.
Similarly, in AndroidControl, Gemini Thinking fails an average of 37 tasks, while Claude Thinking fails 16 tasks.
These results indicate that the reasoning process in current VLMs significantly reduces accuracy on tasks that they could otherwise complete without reasoning.

Second, for tasks that are impossible to complete by non-reasoning models, reasoning provides a moderate improvement.
For instance, Gemini Thinking achieves average improvements of 7.7\%, 8.1\%, and 8.8\% across the three benchmarks, respectively.
However, in most cases--except for Claude Thinking, which shows significant improvement in AndroidWorld with M3A-SoM--these accuracy gains do not compensate for the overall reduction in task completion rates caused by the reasoning process.
Thus, existing reasoning VLMs may have a slightly negative impact on mobile GUI agents.

\input{fig-screenspot-error}

\subsection{Error Analysis}
\label{sec:err-analysis}

We then take a deeper look at the tasks where reasoning VLMs lead to shifts from \textit{completed} to \textit{failed}, aiming to contrast reasoning and non-reasoning VLMs.
In ScreenSpot, we find that approximately all errors are attributed to incorrect grounding coordinate outputs, as demonstrated in Figure~\ref{fig:error-grounding}.
This suggests a significant limitation in the grounding capability of the current VLM reasoning process, which is a key functionality required by mobile GUI agents.

However, in another static benchmark, AndroidControl, grounding errors nearly disappear due to a better mobile GUI agent design.
By incorporating view hierarchies (which include bounding boxes for each GUI element) alongside screenshots as input, mobile GUI agents can more precisely extract the coordinates of GUI elements during reasoning.
Nevertheless, we also observe a large number of errors causing mobile GUI agents to fail in previously successful tasks under non-reasoning modes.

\input{tab-error-analysis}

We combine all tasks with T$\rightarrow$F under all setups using Claude and Claude Thinking, manually identify the errors, and then categorize them based on their sources: \textit{benchmark} and \textit{VLM}.
We present different errors within each category across a total of 67 tasks, along with their explanations and percentage distributions, in Table~\ref{tab:error-analysis}.
All tasks were executed on Claude models, as the API provides comprehensive reasoning processes for our diagnosis, whereas Gemini Thinking's API does not yet support this functionality.

$\bullet$ \textbf{Benchmark} contributes to more than 70\% errors.
The most significant portion of errors (47.8\%) stems from the ``Weak Evaluation Method'', where various correct actions that could continue or complete a task are evaluated as incorrect.
This is a common limitation of static benchmarks and has been noted in previous studies~\cite{zhang2024llamatouch,rawles2024androidworld}.
Another major issue (14.9\%) is the ``Static GUI Input Limitation''.
Since the benchmark feeds only one GUI at a time, the reasoning VLM struggles to determine whether prior states satisfy the requirements of a given task instruction.
After reasoning, it may attempt to revert and check whether the prior condition was met, leading to incorrect outputs compared to the benchmark.
Additionally, some task instructions within the benchmark are unclear, making them difficult for even humans to understand, and thus unsuitable for mobile GUI agents.

$\bullet$ \textbf{VLM.}
The remaining 25.4\% of errors stem from limitations in current reasoning VLMs.
The most significant one is ``Limited GUI Comprehension'',  where during the reasoning phase, the VLM misinterprets the GUI context and generates incorrect responses.
More critically, even if the VLM deduces the correct output during reasoning, it may produce an inconsistent final response.
These inconsistencies further downgrade the performance.
Additionally, we observe a few grounding errors, reasoning errors, and hallucinations after applying reasoning.
Demonstrations of these errors can be found in Appendix~\ref{appendix:errs}.

\input{fig-token-analysis}
\subsection{Token Costs}
\label{sec:token-cost}

Another concern regarding the integration of reasoning VLMs in mobile GUI agents is their high latency and substantial token costs during the reasoning process.
To quantitatively assess this issue, we calculate and compare the number of model output tokens across all benchmarks and agent setups, both with and without reasoning enabled.
We focus particularly on the Claude models, as they explicitly expose their reasoning process.
For Claude Thinking, we accumulate the number of tokens generated during both the reasoning process and the final responses.
The results in Figure~\ref{fig:token-analysis} show that across all three benchmarks and setups, the reasoning process incurs at least 3.11$\times$ the token cost, with a maximum increase of 14.78$\times$.
More specifically, on ScreenSpot, the average number of output tokens without reasoning is 37.6, whereas enabling reasoning increases this value to 238.5.
This significantly raises both token costs and response time, although prior results indicate no considerable performance improvements.
These findings highlight an important question: when should mobile GUI agents leverage advanced reasoning VLMs to enhance performance while maintaining acceptable latency and monetary costs?
Another observation is that on AndroidControl and AndroidWorld, the number of final response tokens remains identical.
However, on ScreenSpot, the reasoning model generates additional information to summarize its reasoning process, resulting in a higher number of response tokens.
This phenomenon stems from weak output constraints in the agent setups.

\subsection{Implications}
\label{sec:implications}
Generally, the reasoning process can provide an in-depth understanding of task instructions and GUIs.
However, this capability does not consistently lead to correct responses when evaluated on static mobile GUI benchmarks due to their intrinsic limitations.
Furthermore, based on the results above, we derive the following implications for improving mobile GUI agent development and evaluation.

$\bullet$ \textit{For VLMs} powering mobile GUI agents:
It is crucial to train VLMs on more comprehensive datasets to enhance their grounding and screen understanding capabilities during reasoning.
This requires large datasets with appropriate annotations~\cite{rawles2023android,li2024effects,gao2024mobileviews}.
Additionally, addressing inconsistencies between reasoning processes and final outputs through robust, domain-specific reward functions in reinforcement learning are essential~\cite{guo2025deepseek}.

$\bullet$ \textit{For benchmarks}: 
Mobile GUI agents should ideally be evaluated on interactive benchmarks due to the inherent limitations of the current evaluation design of static benchmarks (i.e., requiring two identical actions).
Real-world mobile environments could provide richer contextual information, therefore enabling mobile GUI agents to conduct more nuanced reasoning.
Regardless of whether benchmarks are static or interactive, it is crucial to define clear and unambiguous tasks.

$\bullet$ \textit{For mobile GUI agents}: To fully leverage the reasoning capability of VLMs, integrating more contextual information--whether through dynamic innovations in external tools~\cite{wu2025agentic,islam2024open,li2025search,blogGeminiFeatures} or by incorporating holistic information into system prompts--may enhance mobile GUI agents.
Otherwise, without relevant contextual information, the reasoning process is prone to generating suboptimal outcomes.
What's more, adopting adaptive reasoning is crucial for mitigating long latency and high token costs, thereby maintaining the practicality of mobile GUI agents in real-world scenarios.

%% file: tab-acc-screenspot.tex
\begin{table}[t]
    \centering
    \scalebox{0.9}{
    \renewcommand{\arraystretch}{1.2}
\begin{tabular}{c|ccc}
\hline
\multirow{2}{*}{\textbf{VLMs}} & \multicolumn{3}{c}{\textbf{Output Format (to ground the GUI element)}} \\ \cline{2-4} 
 &
  \textbf{\begin{tabular}[c]{@{}c@{}}Normalized\\ Bounding Box\end{tabular}} &
  \textbf{\begin{tabular}[c]{@{}c@{}}Pixel-based\\ Bounding Box\end{tabular}} &
  \textbf{\begin{tabular}[c]{@{}c@{}}Normalized\\ Center Point\end{tabular}} \\ \hline
GPT-4o                         & 33.5\%        & 4.4\%                 & 27.7\%                \\ \hline
Gemini 2.0 Flash               & 50.2\%        & 14.9\%        & 53.0\%       \\ 
Gemini 2.0 Flash Thinking      & 21.5\% \decr{28.7\%}       & 13.6\% \decr{1.3\%}       & 23.3\% \decr{29.7\%}      \\ \hline
Claude 3.7 Sonnet              & 27.5\%        & 2.8\%                 & 6.4\%                 \\
Claude 3.7 Sonnet Thinking     & 11.4\% \decr{16.1\%}       & 2.8\% (-)                & 15.0\% \incr{8.6\%} \\ \hline
\end{tabular}
    \footnotesize
    }
    \vspace{5pt}
    \caption{Mobile GUI grounding accuracy of different VLMs/prompt designs on ScreenSpot~\cite{cheng2024seeclick}.}
    \label{tab:screenspot}
\end{table}

%% file: tab-acc-androidcontrol.tex
\begin{table}[t]
    \centering
    \scalebox{0.85}{
    \renewcommand{\arraystretch}{1.2}
\begin{tabular}{c|cccc}
\hline
\multirow{2}{*}{\textbf{VLMs}} & \multicolumn{4}{c}{\textbf{Agent Designs}}                               \\ \cline{2-5} 
 &
  \textbf{Task Inst.} &
  \textbf{Step Inst.} &
  \textbf{\begin{tabular}[c]{@{}c@{}}Task Inst. + \\ Step Inst.\end{tabular}} &
  \textbf{\begin{tabular}[c]{@{}c@{}}Task Inst. + \\ Prev. Action List\end{tabular}} \\ \hline
GPT-4o                         & 39.2\%               & 66.4\%               & 68\%                & 44.8\%      \\ \hline
Gemini 2.0 Flash               & 40.8\%               & 64.8\%               & 62.8\%              & 46\%  \\ 
Gemini 2.0 Flash Thinking      & 42.6\% \incr{1.8\%}  & 65.6\% \incr{0.8\%}  & 63.6\% \incr{0.8\%} & 45.6\% \decr{0.4\%} \\  \hline
Claude 3.7 Sonnet              & 42.4\%               & 59.4\%              & 58.2\%              & 43\%   \\
Claude 3.7 Sonnet Thinking     & 43.6\% \incr{1.2\%}  & 63.8\% \incr{4.4\%}  & 60.8\% \incr{2.6\%}   & 44\% \incr{1\%} \\ \hline
\end{tabular}
    \footnotesize
    }
    \vspace{5pt}
    \caption{Action prediction accuracy of different VLMs/agent designs on AndroidControl~\cite{li2024effects}.}
    \label{tab:android-control}
\end{table}

%% file: tab-acc-androidworld.tex
\begin{table}[h]
    \centering
    \scalebox{0.9}{
    \renewcommand{\arraystretch}{1.2}
\begin{tabular}{c|ccc}
\hline
\multirow{3}{*}{\textbf{VLMs}} & \multicolumn{3}{c}{\textbf{Agent Designs}}                                                                                                                                                                                                                \\ \cline{2-4} 
                               & \multirow{2}{*}{\textbf{\begin{tabular}[c]{@{}c@{}}M3A\\ (SoM)\end{tabular}}} & \multirow{2}{*}{\textbf{\begin{tabular}[c]{@{}c@{}}M3A\\ (a11y tree)\end{tabular}}} & \multirow{2}{*}{\textbf{\begin{tabular}[c]{@{}c@{}}T3A\\ (a11y tree)\end{tabular}}} \\
                               &                                                                               &                                                                                     &                                                                                     \\ \hline
GPT-4o                         & 44.8\%                                                                        & 23.3\%                                                                              & 46.6\%                                                                              \\ \hline
Gemini 2.0 Flash               & 35.3\%                                                                        & 25.9\%                                                                              & 39.7\%                                                                              \\
Gemini 2.0 Flash Thinking      & 32.8\% \decr{2.5\%}                                                           & 23.2\% \decr{2.7\%}                                                                 & 36.2\% \decr{3.5\%}                                                                 \\ \hline
Claude 3.7 Sonnet              & 55.2\%                                                                        & 44.8\%                                                                              & 54.3\%                                                                              \\
Claude 3.7 Sonnet Thinking     & 64.7\% \incr{9.5\%}                                                             & 50\% \incr{5.2\%}                                                                 & 58.6\% \incr{4.3\%}                                                                          \\ \hline
\end{tabular}
}
\footnotesize
\vspace{5pt}
\caption{Task completion rates with different agent designs on AndroidWorld~\cite{rawles2024androidworld}.}
\label{tab:android-world}
\end{table}

%% file: fig-androidworld-difficulty.tex
\begin{figure}[h]
	\centering					
	\includegraphics[width=1\textwidth]{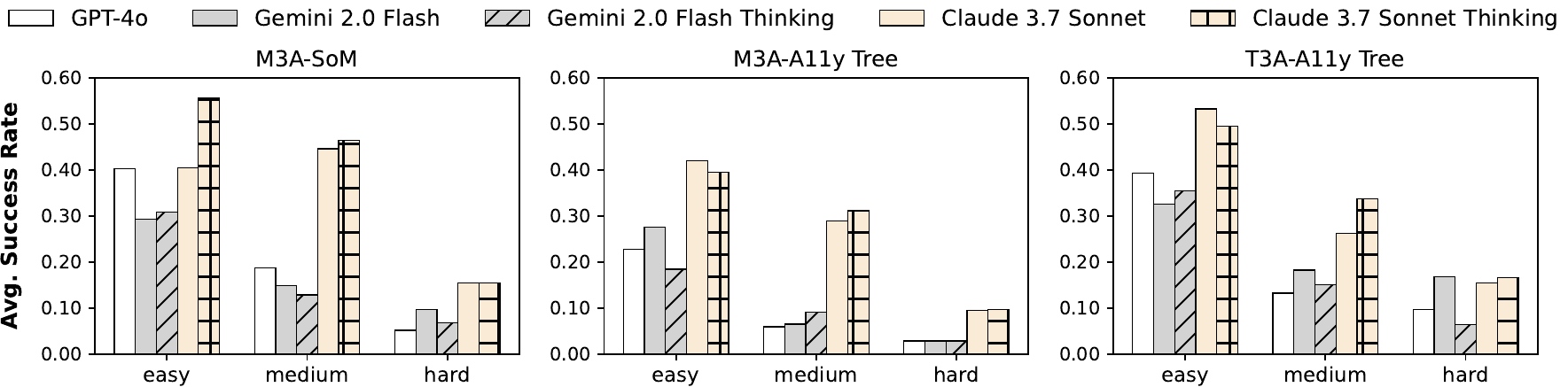}
	\caption{Task completion rates on AndroidWorld categorized by task difficulties.}
	\label{fig:androidworld-difficulty}
\end{figure}

%% file: tab-task-stats.tex
\begin{table}[h]
\centering
    \scalebox{0.8}{
    \renewcommand{\arraystretch}{1.2}
\begin{tabular}{c|c|cccc|cccc}
\hline
\multirow{2}{*}{\textbf{Benchmark}} &
  \multirow{2}{*}{\textbf{Setup}} &
  \multicolumn{4}{c|}{\textbf{Gemini 2.0 Flash}} &
  \multicolumn{4}{c}{\textbf{Claude 3.7 Sonnet}} \\ \cline{3-10} 
 &
   &
  \textit{\textbf{T$\rightarrow$F}} &
  \textit{\textbf{F$\rightarrow$T}} &
  \textit{\textbf{T$\rightarrow$T}} &
  \textit{\textbf{F$\rightarrow$F}} &
  \textit{\textbf{T$\rightarrow$F}} &
  \textit{\textbf{F$\rightarrow$T}} &
  \textit{\textbf{T$\rightarrow$T}} &
  \textit{\textbf{F$\rightarrow$F}} \\ \hline
\multirow{3}{*}{ScreenSpot}     & Norm. BBox          & 36.06\% & 7.37\%  & 14.14\% & 42.43\% & 17.93\% & 1.79\%  & 9.56\%  & 70.72\% \\
                                & Pixel BBox         & 11.55\% & 10.16\% & 3.39\%  & 74.90\% & 0.40\%  & 0.40\%  & 2.39\%  & 96.81\% \\
                                & Center Point        & 35.26\% & 5.58\%  & 17.73\% & 41.43\% & 0.99\%  & 9.56\%  & 5.38\%  & 84.06\% \\ \hline
\multirow{4}{*}{AndroidControl} & Task Inst.    & 8.42\%  & 10.22\% & 32.46\% & 48.9\%  & 3.4\%   & 4.6\%   & 39.0\%  & 53.0\%  \\
                                & Step Inst.    & 5.0\%   & 5.8\%   & 59.8\%  & 29.4\%  & 1.0\%   & 5.4\%   & 58.4\%  & 35.2\%  \\
                                & Task Inst. + Step   & 6.8\%   & 7.6\%   & 56.0\%  & 29.6\%  & 3.2\%   & 5.8\%   & 55.0\%  & 36.0\%  \\
                                & Task Inst. + Act.   & 9.2\%   & 8.8\%   & 36.8\%  & 45.2\%  & 5.8\%   & 6.8\%   & 37.2\%  & 50.2\%  \\ \hline
\multirow{3}{*}{AndroidWorld} & M3A-SoM       & 12.17\% & 10.43\% & 22.61\% & 54.78\% &   0.86\%  & 10.43\% & 54.31\% & 34.48\% \\
                                & M3A-A11y Tree & 10.43\% & 6.09\%  & 15.56\% & 67.83\% & 6.03\% & 11.21\%  & 38.79\% & 43.97\% \\
                                & T3A-A11y Tree & 12.28\% & 9.65\%  & 27.19\% & 50.88\% & 5.17\%  & 9.48\%  & 49.14\% & 36.21\% \\ \hline
\end{tabular}
    }
    \vspace{5pt}
    \caption{Task completion statistics (\% of all tasks) across benchmarks and task setups with reasoning and non-reasoning VLMs.
    \textbf{T$\rightarrow$F}: Tasks completed in non-reasoning mode but failed in reasoning mode; \textbf{F$\rightarrow$T}: Tasks failed in non-reasoning mode but completed in reasoning mode; \textbf{T$\rightarrow$T}: Tasks completed in both modes; \textbf{F$\rightarrow$F}: Tasks failed in both modes. A high T$\rightarrow$F value indicates a negative impact of the reasoning process; a high F$\rightarrow$T value indicates a positive impact.}
    \label{tab:task-stats}
    \end{table}

%% file: fig-screenspot-error.tex
\begin{figure}[t]
\centering
\includegraphics[width=0.98\textwidth]{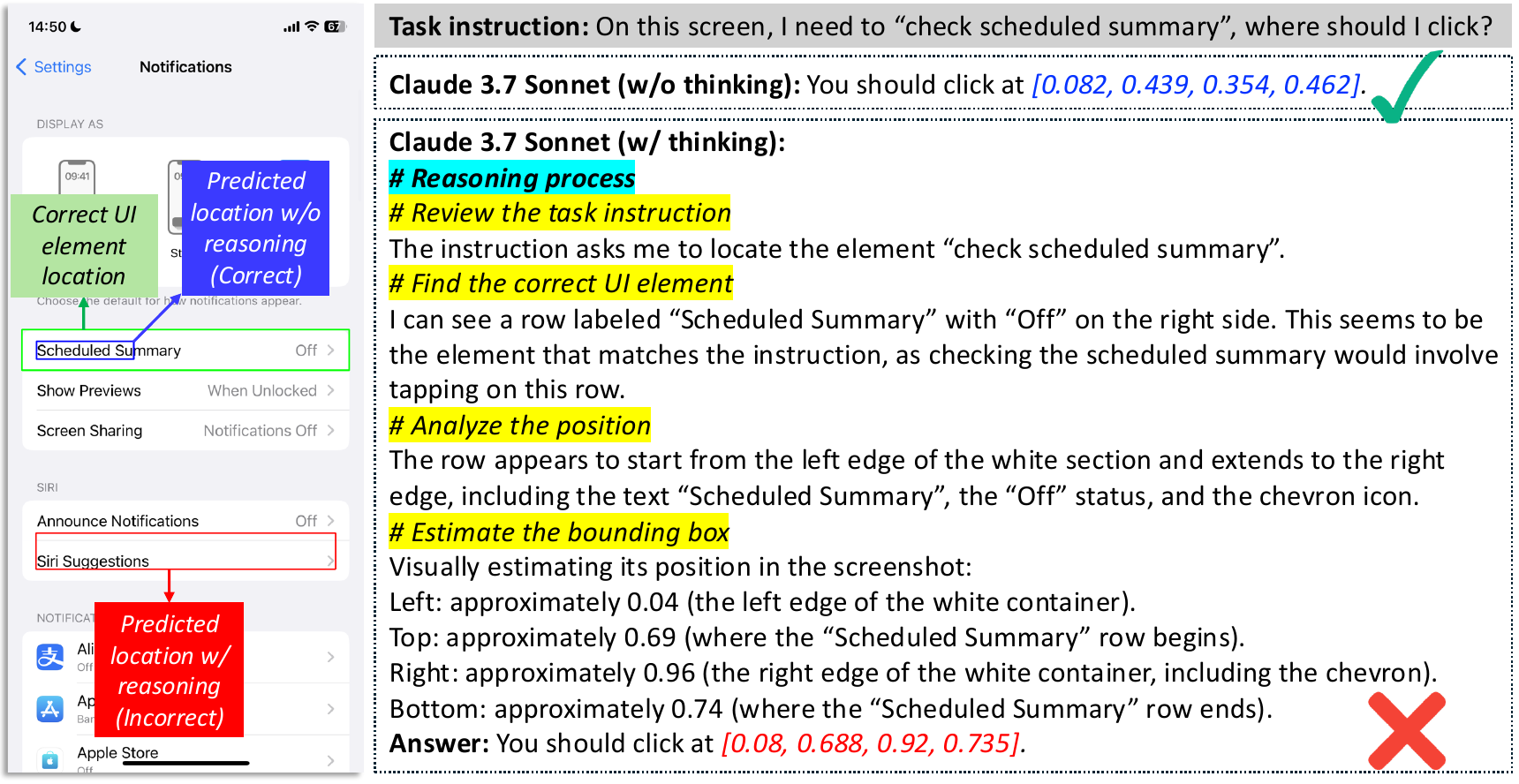}
\caption{An example of a grounding error on ScreenSpot.}
\label{fig:error-grounding}
\end{figure}

%% file: tab-error-analysis.tex
\begin{table}[h]
\centering
\scalebox{0.9}{
\renewcommand{\arraystretch}{1.15}
\begin{tabular}{c|c|c|c|c}
\hline
\textbf{Error Source} &
  \textbf{Error Type} &
  \textbf{Explanation} &
  \textbf{Percentage} &
  \multicolumn{1}{c}{\textbf{Example}} \\ \hline
\multirow{3}{*}{\textit{Benchmark}} &
  \begin{tabular}[c]{@{}c@{}}Weak Evaluation\\ Method\end{tabular} &
  \begin{tabular}[c]{@{}c@{}}Various false negative actions\\ may complete a task\end{tabular} &
  47.8\% &
  Fig.~\ref{fig:error-weak-evaluation-method}
   \\ \cline{2-5} 
 &
  \begin{tabular}[c]{@{}c@{}}Static GUI Input\\ Limitation\end{tabular} &
  \begin{tabular}[c]{@{}c@{}}GUI agents receive only\\ static, individual mobile GUIs\end{tabular} &
  14.9\% &
  Fig.~\ref{fig:error-static-GUI-input-limitation}
   \\ \cline{2-5} 
 &
  \begin{tabular}[c]{@{}c@{}}Unclear Task\\ Instruction\end{tabular} &
  \begin{tabular}[c]{@{}c@{}}Vague or ambiguous task \\ instructions\end{tabular} &
  11.9\% &
  Fig.~\ref{fig:error-unclear-task-instruction}
   \\ \hline
\multirow{3}{*}{\textit{VLM}} &
  \begin{tabular}[c]{@{}c@{}}Limited GUI\\ Comprehension\end{tabular} &
  \begin{tabular}[c]{@{}c@{}}Unable to fully understand \\ the GUI context\end{tabular} &
  10.5\% &
  Fig.~\ref{fig:error-limited-GUI-comprehension}
   \\ \cline{2-5} 
 &
  \begin{tabular}[c]{@{}c@{}}Reasoning-Response\\ Inconsistency\end{tabular} &
  \begin{tabular}[c]{@{}c@{}}Correct reasoning process\\ but inconsistent response\end{tabular} &
  8.9\% &
  Fig.~\ref{fig:error-reasoning-output-inconsistency}
   \\ \cline{2-5} 
 &
  Others &
  \begin{tabular}[c]{@{}c@{}}Incorrect grounding, incorrect \\reasoning, and hallucination\end{tabular} &
  6.0\% &
  Fig.~\ref{fig:error-grounding-error-androidcontrol}/~\ref{fig:error-reasoning-error}/~\ref{fig:error-hallucination}
   \\ \hline
\end{tabular}
	}
	\footnotesize
	\vspace{5pt}
	\caption{AndroidControl error analysis for tasks completed by Claude without reasoning but failed with reasoning enabled (i.e., \textbf{T$\rightarrow$F} in Table~\ref{tab:task-stats}).
    Examples of each error are illustrated in Appendix~\ref{appendix:errs}.}
	\label{tab:error-analysis}
	\end{table}

%% file: fig-token-analysis.tex
\begin{figure}[h]
	\centering
	\includegraphics[width=0.99\textwidth]{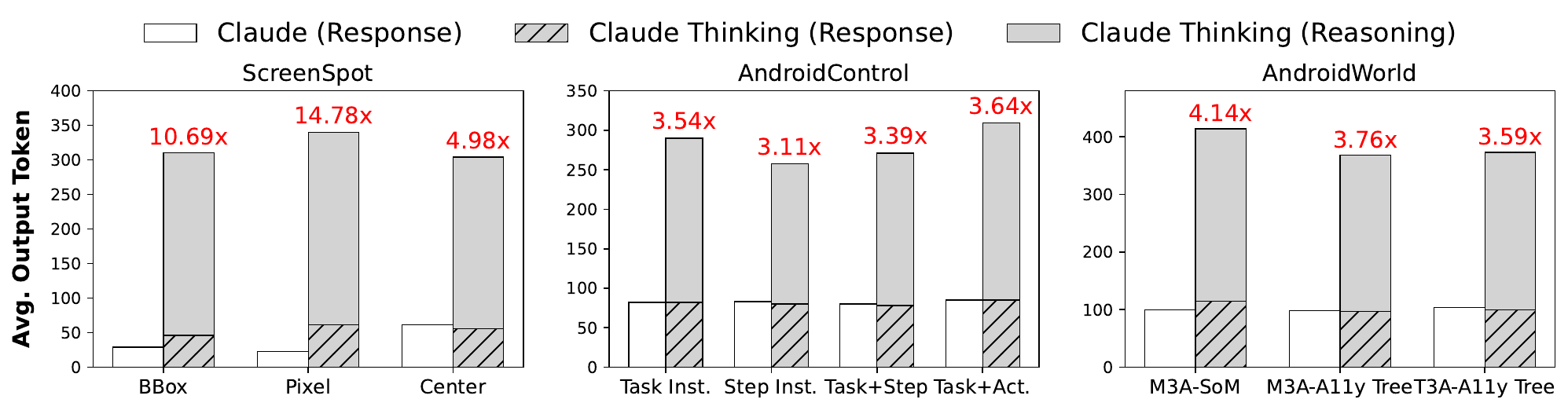}
	\caption{Comparison of average output token count between the Claude reasoning model and its base model without reasoning. Across all setups, reasoning increases token consumption by at least 3$\times$ compared to the non-reasoning model, resulting in higher monetary costs and increased response latency.}
	\label{fig:token-analysis}
\end{figure}

%% file: conclusion.tex
\section{Conclusions and Future Work}
\label{sec:conclusion}

In this work, we conduct the first empirical study to investigate whether the reasoning capabilities of commercial VLMs enhance the performance of mobile GUI agents.
Using two series of commercial VLMs (i.e., Gemini 2.0 Flash and Claude 3.7 Sonnet) with and without reasoning enabled, we comprehensively evaluate various mobile GUI agents under different configurations across two static benchmarks (i.e., ScreenSpot and AndroidControl) and one interactive benchmark (i.e., AndroidWorld).
We report the overall trend in task completion rates across the three benchmarks and provide a deeper analysis on a per-task basis.
Although we observe SOTA performance on the AndroidWorld benchmark, current reasoning-enabled VLMs generally provide only marginal or even negative improvements in mobile GUI agent performance, with a significant concern that they often fail tasks that could be completed without reasoning.
We categorize the errors arising from the reasoning process and offer practical guidance for future research on improving mobile GUI agents, VLMs, and benchmarks.

As the next step, we will explore additional benchmarks (e.g., Desktop tasks \cite{xie2024osworld}), models (e.g., open-source or small language models \cite{lu2024small}), and agentic workflows (e.g., external tool-enabled approaches \cite{li2024personal}) to comprehensively evaluate the effectiveness of CoT reasoning in GUI tasks.

%% file: appendix.tex
\section{Appendix}

\subsection{Demonstrations of Errors by Reasoning VLMs in AndroidControl}
\label{appendix:errs}

Table~\ref{tab:error-analysis} categorizes the errors introduced by reasoning VLMs in AndroidControl.
In this section, we illustrate these errors using traces collected during our experiments.
We present the inputs and responses of both non-reasoning and reasoning VLMs.

\input{fig-androidcontrol-error}

%% file: fig-androidcontrol-error.tex
\begin{figure}[h!]
\centering
\includegraphics[width=0.9\textwidth]{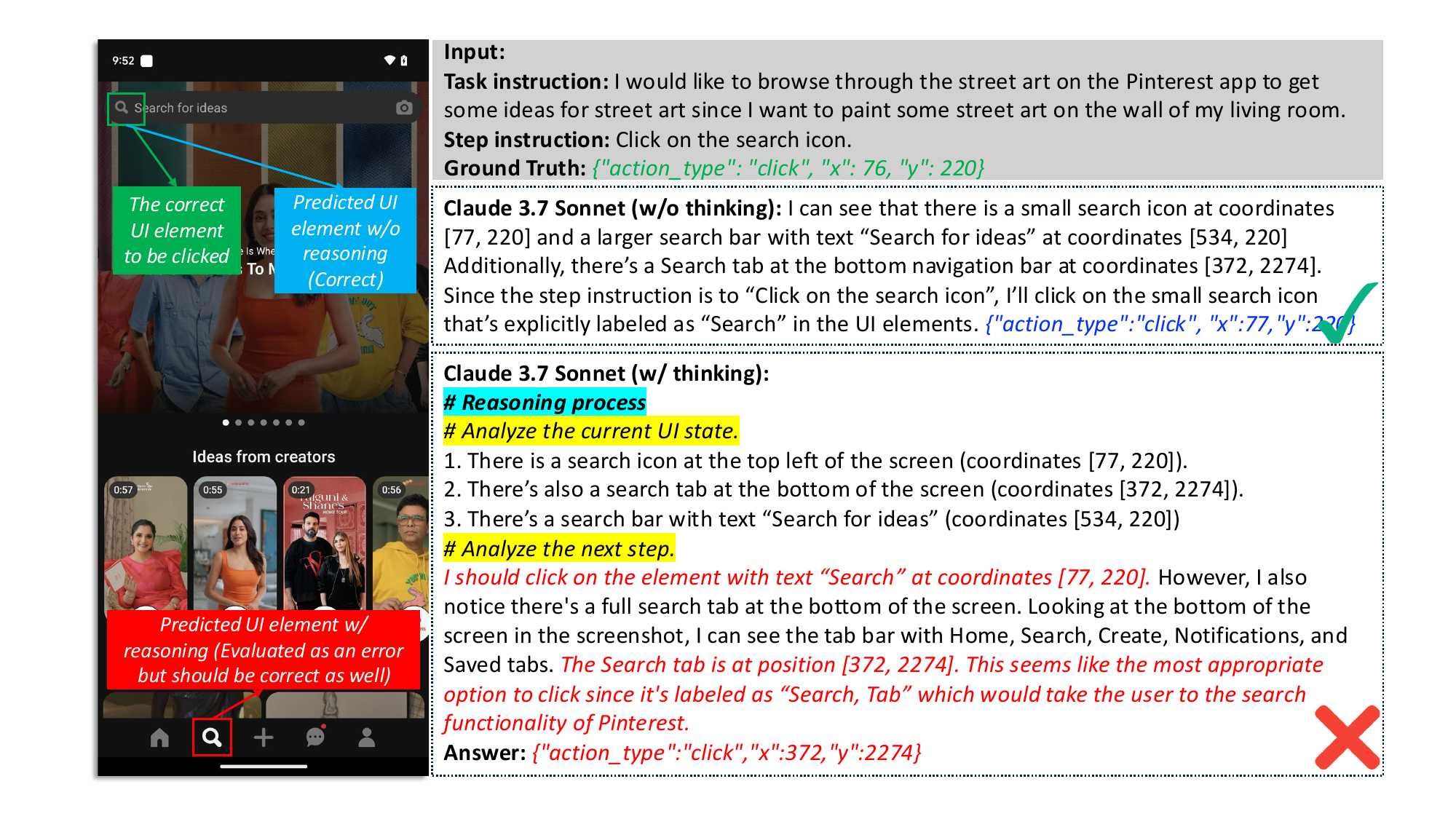}
\caption{Benchmark Error: Weak Evaluation Method.}
\label{fig:error-weak-evaluation-method}
\end{figure}

\begin{figure}[h!]
\centering
\includegraphics[width=0.9\textwidth]{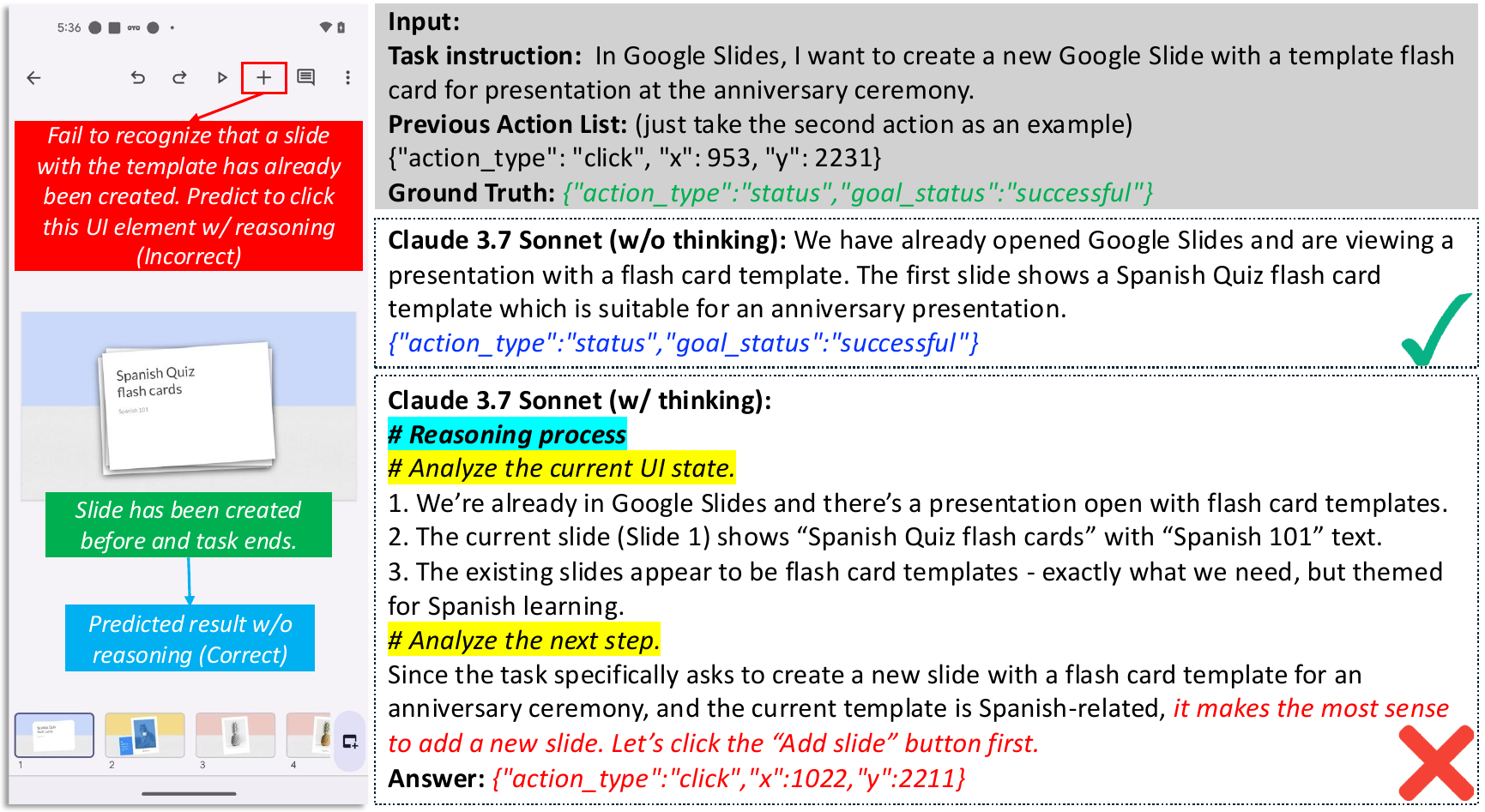}
\caption{Benchmark Error: Static GUI Input Limitation.}
\label{fig:error-static-GUI-input-limitation}
\end{figure}

\begin{figure}[h!]
\centering
\includegraphics[width=0.9\textwidth]{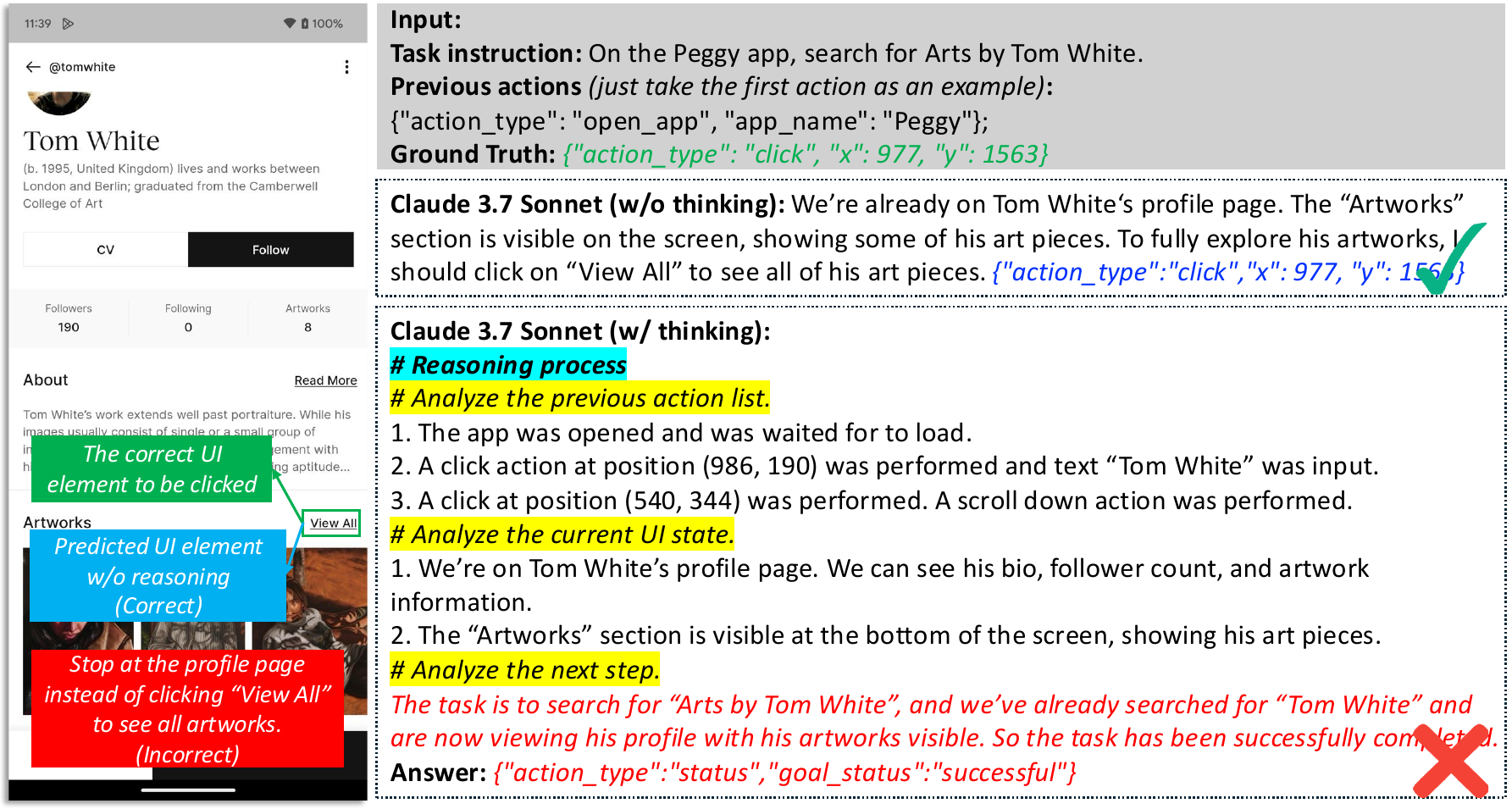}
\caption{Benchmark Error: Unclear Task Instruction.}
\label{fig:error-unclear-task-instruction}
\end{figure}

\begin{figure}[h!]
\centering
\includegraphics[width=0.9\textwidth]{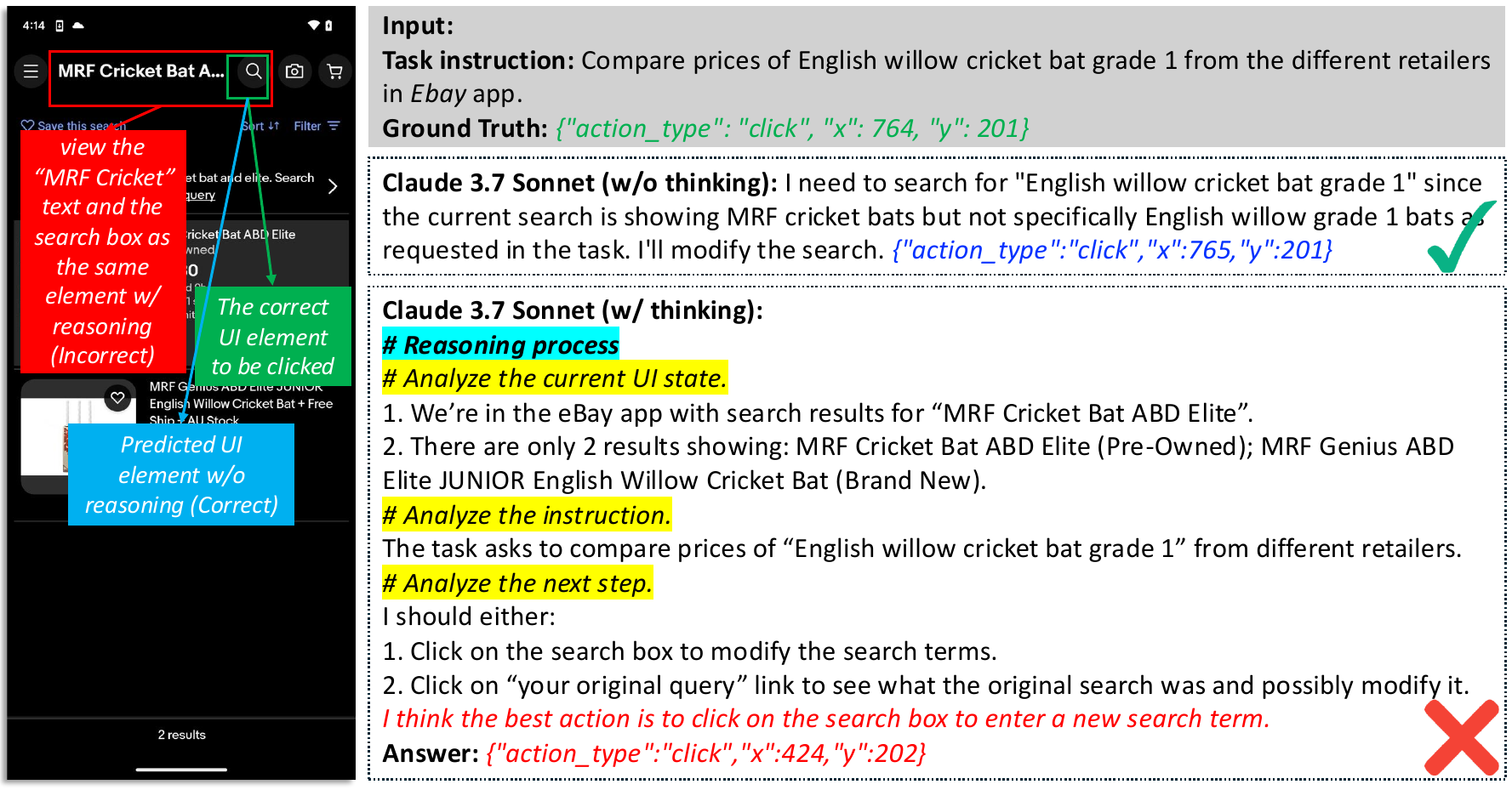}
\caption{VLM Error: Limited GUI Comprehension.}
\label{fig:error-limited-GUI-comprehension}
\end{figure}

\begin{figure}[h!]
\centering
\includegraphics[width=0.9\textwidth]{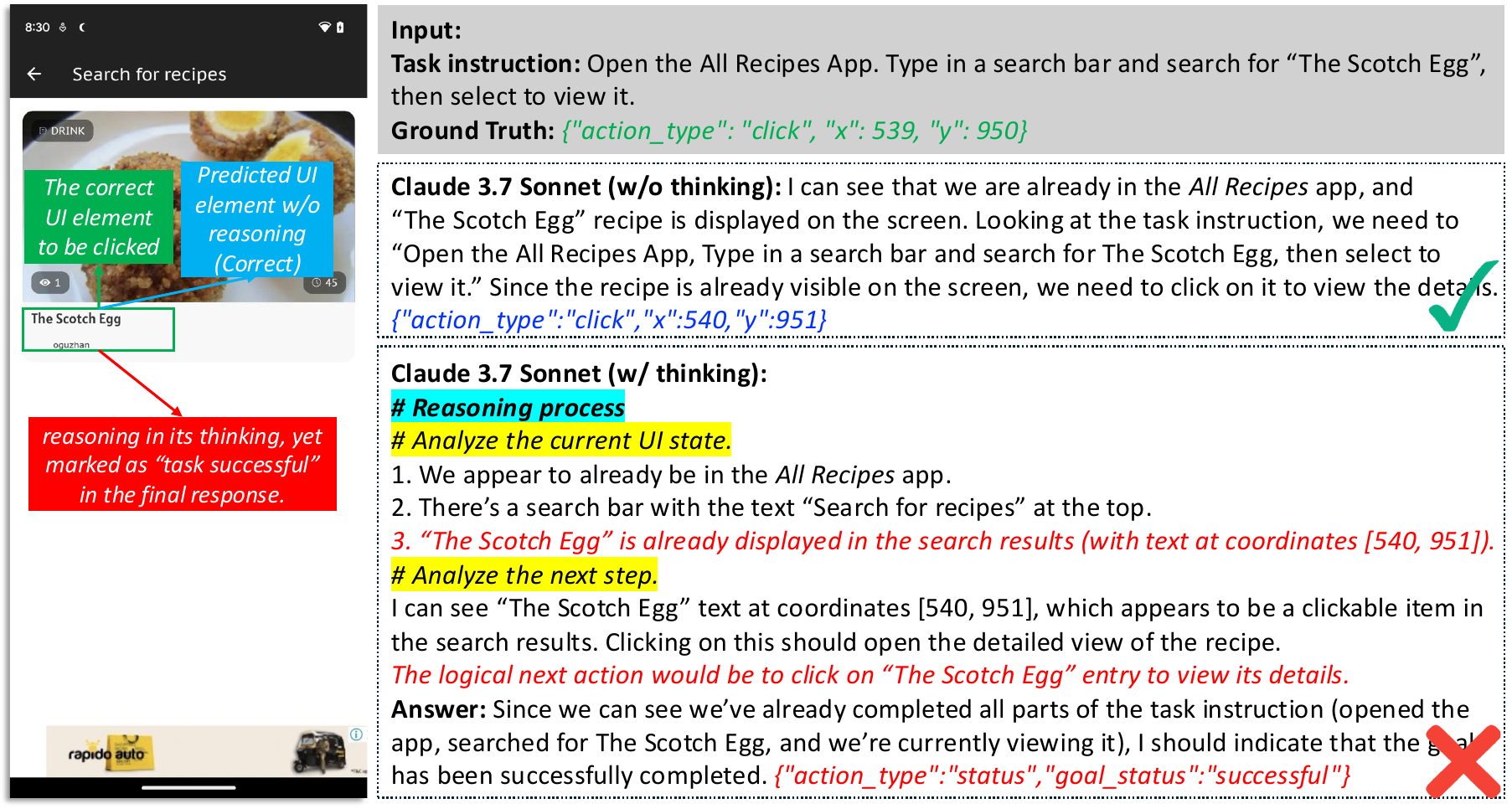}
\caption{VLM Error: Reasoning-Response Inconsistency.}
\label{fig:error-reasoning-output-inconsistency}
\end{figure}

\begin{figure}[h!]
\centering
\includegraphics[width=0.9\textwidth]{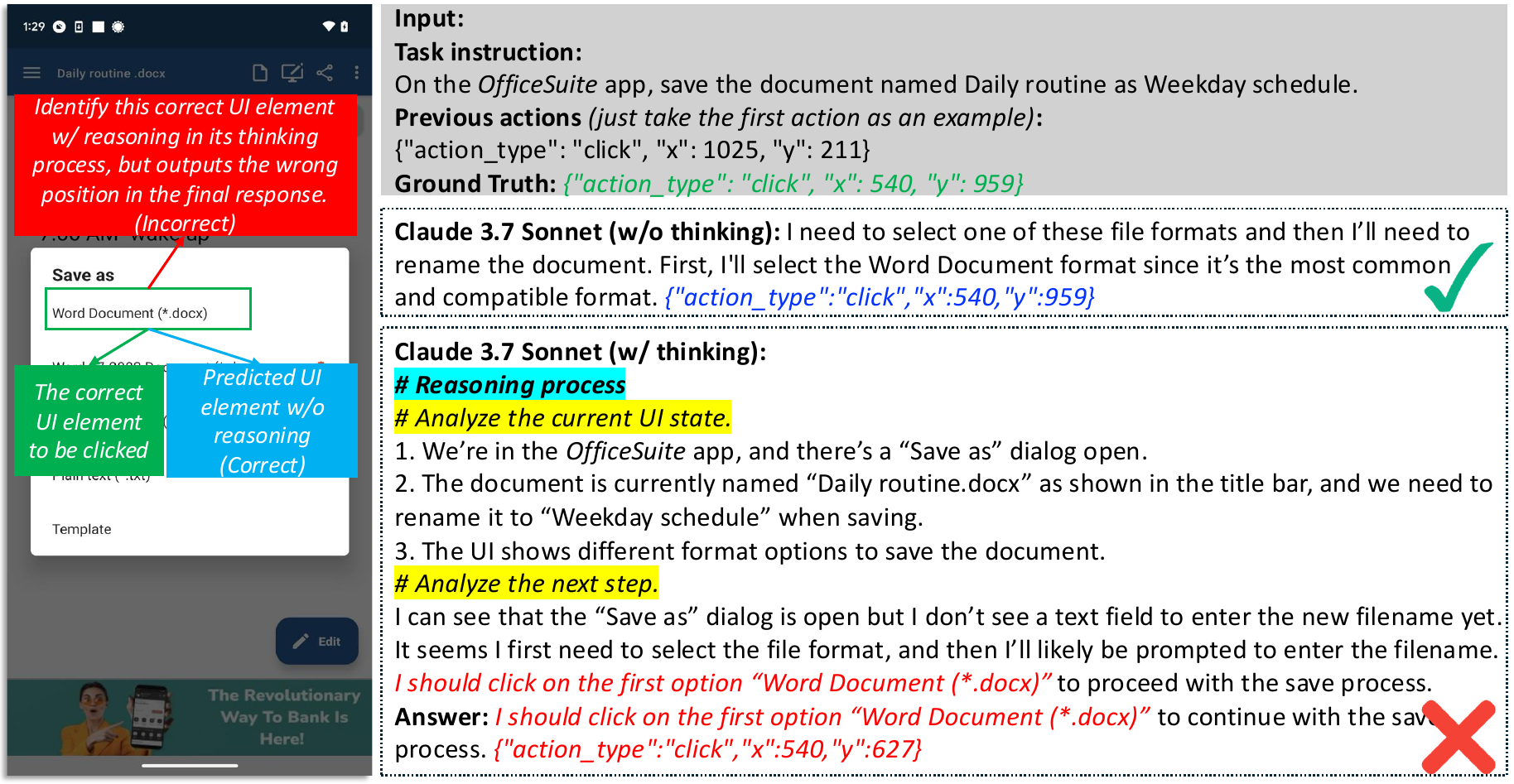}
\caption{VLM Error: Incorrect Grounding.}
\label{fig:error-grounding-error-androidcontrol}
\end{figure}

\begin{figure}[h!]
\centering
\includegraphics[width=0.9\textwidth]{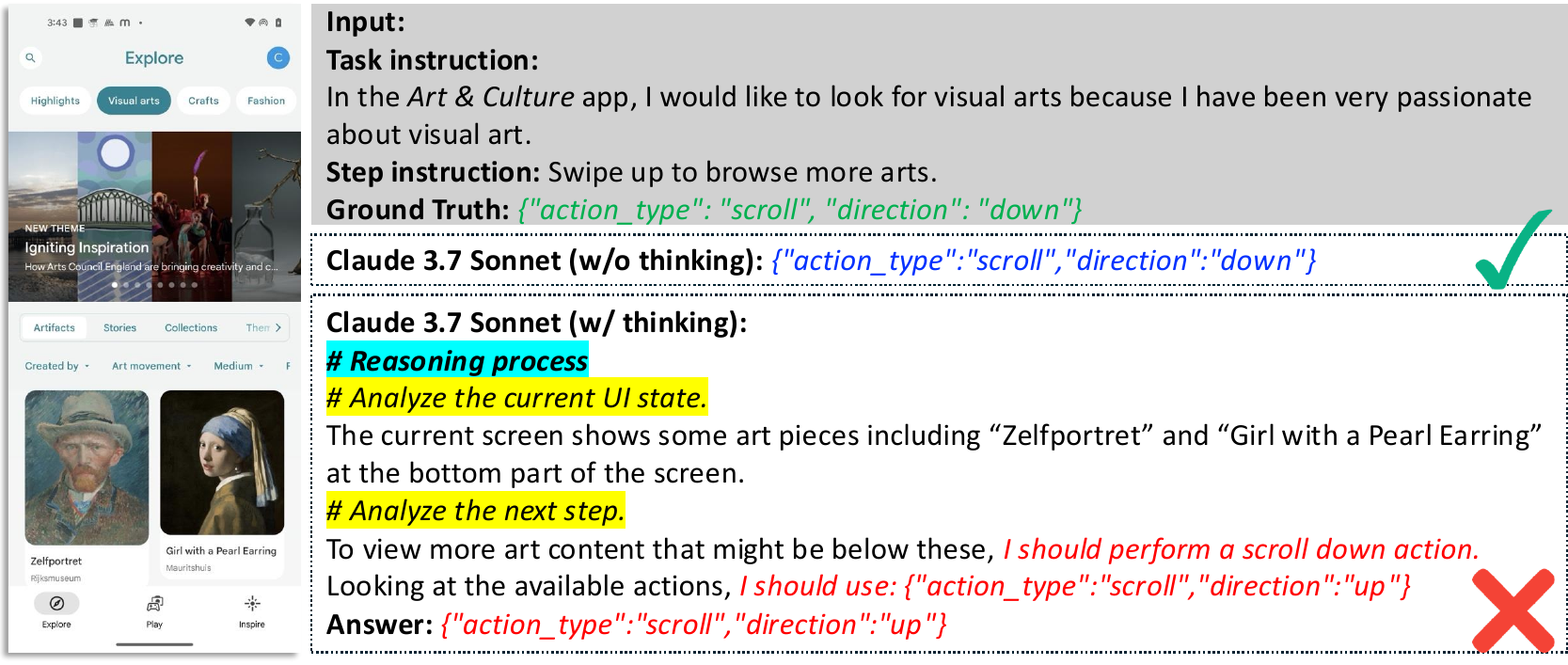}
\caption{VLM Error: Incorrect Reasoning. The model correctly infers that it should scroll down to view more content but incorrectly outputs a scroll-up action in both its reasoning process and final answer.}
\label{fig:error-reasoning-error}
\end{figure}

\begin{figure}[h!]
\centering
\includegraphics[width=0.9\textwidth]{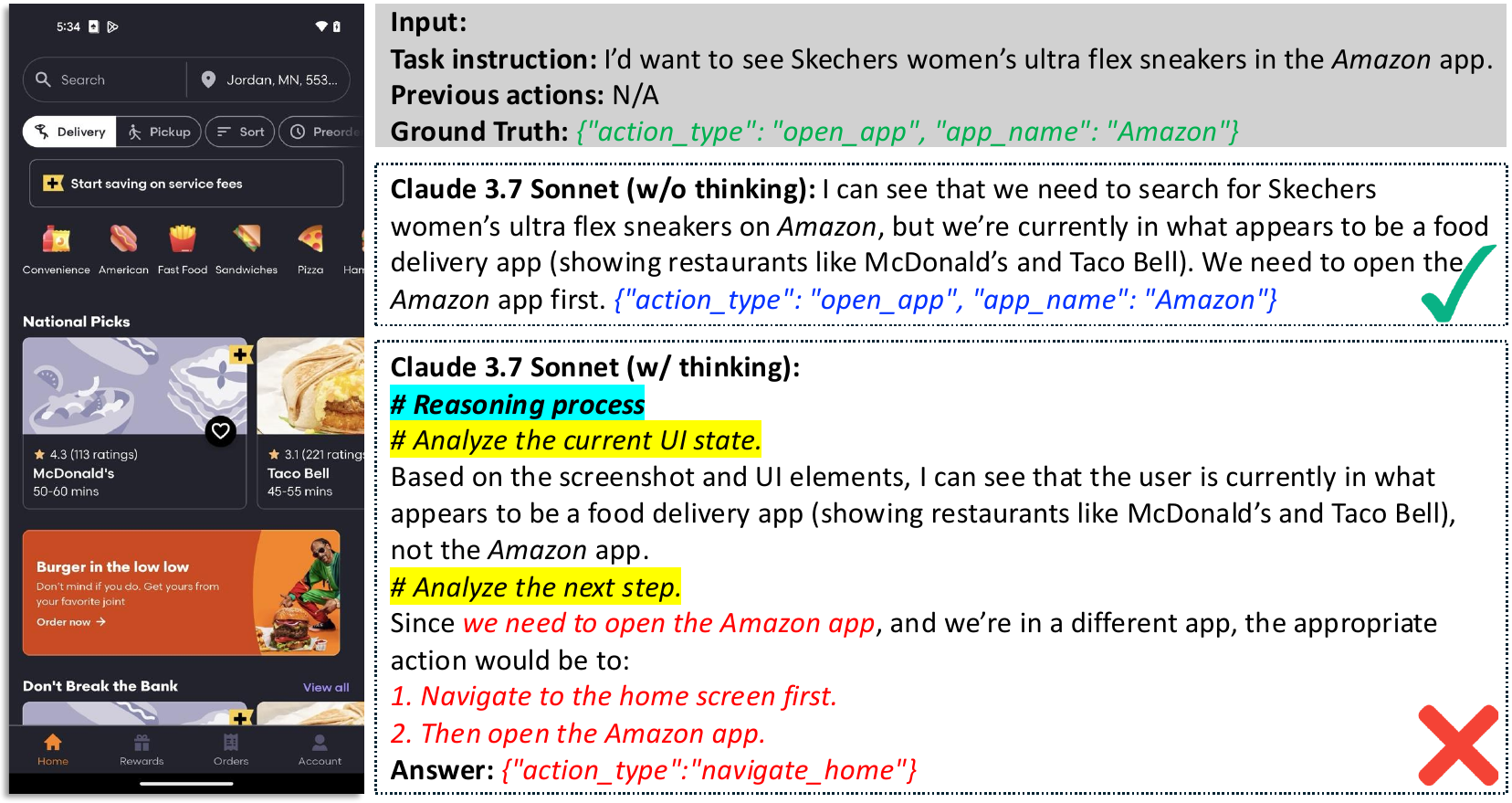}
\caption{VLM Error: Hallucination. The ``navigate\_home'' action is not in the given action space.}
\label{fig:error-hallucination}
\end{figure}